%% file: main.tex
\documentclass{article}

\usepackage{arxiv}

\usepackage[utf8]{inputenc} % allow utf-8 input
\usepackage[T1]{fontenc}    % use 8-bit T1 fonts
\usepackage{hyperref}       % hyperlinks
\usepackage{booktabs}       % professional-quality tables
\usepackage{amsfonts}       % blackboard math symbols
\usepackage{nicefrac}       % compact symbols for 1/2, etc.
\usepackage{microtype}      % microtypography
\usepackage{graphicx}
\usepackage{enumitem}
\usepackage{amsmath}
\usepackage{xcolor}
\usepackage{dsfont}
\usepackage{bbm}
\usepackage{tikz}
\usepackage{tikz-qtree}
\usepackage{tikz-qtree-compat}
\usepackage{url}

\usepackage{array}
\newcolumntype{C}[1]{>{\centering\arraybackslash\hspace{0pt}}m{#1}}

\DeclareMathOperator*{\argmax}{arg\,max}
\DeclareMathOperator*{\argmin}{arg\,min}

\title{Tree-Constrained Graph Neural Networks\\for Argument Mining}

\author{ \href{https://orcid.org/0000-0002-1697-8586}{\includegraphics[scale=0.06]{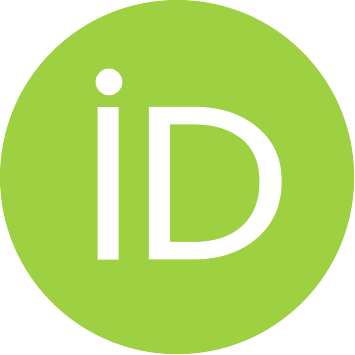}\hspace{1mm}Federico Ruggeri} \\
	DISI \\
	University of Bologna \\
	Bologna, Italy \\
	\texttt{federico.ruggeri6@unibo.it} \\
	\And
	\href{https://orcid.org/0000-0002-9663-1071}{\includegraphics[scale=0.06]{orcid.pdf}\hspace{1mm}Marco Lippi} \\
	DISMI\\
	University of Modena and Reggio Emilia\\
	Reggio Emilia, Italia \\
	\texttt{marco.lippi@unimore.it} \\
    \And 
    \href{https://orcid.org/0000-0002-9253-8638}{\includegraphics[scale=0.06]{orcid.pdf}\hspace{1mm}Paolo Torroni} \\
    DISI \\
    University of Bologna \\
	Bologna, Italy \\
	\texttt{p.torroni@unibo.it} \\
}

\date{}

\renewcommand{\shorttitle}{Tree-Constrained Graph Neural Networks for Argument Mining}

\hypersetup{
pdftitle={Tree-Constrained Graph Neural Networks\\for Argument Mining},
pdfsubject={cs.AI},
pdfauthor={Federico Ruggeri, Marco Lippi, Paolo Torroni},
pdfkeywords={Tree kernels, graph neural networks, argument mining, differentiable pooling},
}

\begin{document}
\maketitle

\begin{abstract}
    We propose a novel architecture for Graph Neural Networks that is inspired by the idea behind Tree Kernels of measuring similarity between trees by taking into account their common substructures, named fragments. By imposing a series of regularization constraints to the learning problem, we exploit a pooling mechanism that incorporates such notion of fragments within the node soft assignment function that produces the embeddings. We present an extensive experimental evaluation on a collection of sentence classification tasks conducted on several argument mining corpora, showing that the proposed approach performs well with respect to state-of-the-art techniques.
\end{abstract}

% keywords can be removed

\section{Introduction} \label{sec:introduction}
\input introduction.tex

\section{Background} \label{sec:background}
\input background.tex

\section{Tree-Constrained GNNs} \label{sec:tc-gnns}
\input tcgnns.tex

\section{Experiments} \label{sec:experiments}
\input experiments.tex

\section{Related Work} \label{sec:related_work}
\input related.tex

\section{Conclusions} \label{sec:conclusions}
\input conclusions.tex

\section*{Acknowledgment}
The authors would like to thank Andrea Galassi and Michele Lombardi for fruitful discussions. Paolo Torroni and Federico Ruggeri have been partially supported by the H2020 Project AI4EU (Grant Agreement No. 825619).

\appendix
\section{GNN architecture calibration} \label{appendix:gnn_architecture}
\input appendix.tex

\bibliographystyle{unsrt}
\bibliography{main}

\end{document}

%% file: introduction.tex
Graph Neural Networks (GNNs) are currently a hot topic in artificial intelligence, with a huge amount of applications in many domains, ranging from bioinformatics to computer vision, from social network analysis to natural language processing~\cite{wu2020comprehensive}. First introduced in~\cite{scarselli2008graph}, and then far and wide extended with a large number of variants, GNNs can learn embedding representations of generic graphs, by exploiting aggregation functions based on propagation and pooling layers. These building blocks are frequently stacked into a deep network, and the resulting embeddings can be exploited in any high-level task. This kind of architecture has rapidly become the state-of-the-art, or at least a strong competitor, in many application domains dealing with structured data.

Historically, in natural language processing (NLP) as well as in other domains, Tree Kernels (TKs) have long been one of the most widely employed technique to handle structured data in the form of trees~\cite{moschitti2006making}. A TK is basically a similarity function that captures the degree of similarity of two trees by looking at common fragments within their substructures. Different definitions of fragments induce different definitions of kernels.
In this paper, we propose a novel architecture for GNNs that is inspired by the core idea behind TKs of looking for relevant fragments within the input tree structure.

Specifically, we modify the GNN learning problem by introducing constraints that enforce a regularization term via node soft assignment rules, which correspond to specific definitions of TK fragments. In this way, the clustering of nodes obtained by the pooling function will encode a set of tree sub-structures matching such fragments.
We remark that our architecture is not specifically designed for trees, but indeed it can handle generic graphs as input. Yet, the focus of this paper, especially in the experimental part, will be on input trees.

As an application domain, we focus on argument mining, whose goal is to automatically extract arguments from unstructured textual corpora. One crucial step in argument mining is to detect sentences containing relevant argument components, such as a claim (the conclusion of an argument) or a premise (an evidence supporting a claim). Recent studies~\cite{lippi2015context,lippi2016margot}, have shown that the structure of a sentence is often highly significant for identifying the presence of an argument, or part thereof. For this reason, TKs operating on constituency trees have been successfully used within this domain.

This kind of approach significantly differs from the extensions of GNNs that have been proposed in the literature, in the last decade, to handle tree structures. Architectures such as Tree-LSTMs~\cite{tai2015} and TreeNet~\cite{cheng2018treenet} are basically designed so as to compute tree representations in a bottom-up fashion, by aggregating in a given node the contributions of its children. Conversely, our approach exploits regulatization constraints inspired by TKs, to drive the training phase in building significant hierarchical clusters.

Our main contributions can be summarized as follows.
\begin{itemize}
    \item We propose a GNN architecture where each node cluster contributing to the embedding is compliant to a specific constraint imposed by a given TK. The problem is framed as a node soft assignment task, where constraints are encoded as regularization terms within a plug-and-play pooling layer based on DiffPool~\cite{ying2018diffpool}.
    \item We exploit a dual Lagrangian framework to learn the coefficients of the constraints, which are dynamically updated throughout the training phase, according to the degree of violations of each constraint.
    \item We perform an extensive experimental evaluation on various datasets for the challenging task of sentence classification in argument mining.
\end{itemize}

The paper is structured as follows: Section~\ref{sec:background} presents some background on GNNs, differentiable pooling, and TKs. Then, Section~\ref{sec:tc-gnns} describes our Tree-Constrained GNN, whereas Section~\ref{sec:experiments} illustrates and discusses the experimental results. Finally, Section~\ref{sec:conclusions} concludes the paper.

%% file: background.tex
\subsection{Graph Neural Networks}

Graph neural networks (GNNs) were first introduced in \cite{scarselli2008graph} and have become a key component for handling structured data~\cite{wu2020comprehensive}. Given their expressive power~\cite{xu2019power}, GNNs have been applied to a wide variety of domains~\cite{wu2020comprehensive}. Several architectures have been proposed, ranging from simple aggregating methods such as Graph Convolutional Network (GCN) \cite{KipfW16} and Message Passing Neural Network (MPNN)~\cite{DuvenaudMAGHAA15, GilmerSRVD17} to advanced architectures like Graph Attention Network (GAT)~\cite{velickovic2018graph} and GraphSAGE~\cite{HamiltonYL17}. Given a graph $\mathcal{G = (\mathcal{V}, \mathcal{E})}$ with $\mathcal{V} = \{v_1, \dots, v_{n}\}$ vertices and $\mathcal{E} = \{ e_1, \dots, e_{m} \}$ edges, a GNN generally employs the following message-passing aggregation function:
\begin{equation}
    H^{t+1} = f(A, H^t; \theta^{t+1})
\end{equation}
where $H^t = \{h^t_1, \dots, h^t_{n} \}$ is the node representation matrix $n \times d$ at time $t$, where $d$ is the embedding dimensionality. Each node $h^t_i$ is the node embedding vector and $A$ is the binary adjacency matrix given $(\mathcal{V}, \mathcal{E})$ and $\theta^t$ are model parameters at time $t$. In this paper, our methodology is based on the GCN architecture due to its simplicity and effectiveness, but the approach is not limited by such a solution. Nonetheless, the proposed setup is applicable to any kind of GNN alike a plug-and-play module. In the case of a GCN, the aggregation function takes the following grounding:
\begin{equation} \label{eq:gcn:message-passing}
    H^{t+1} = ReLU(\tilde{A} H^t W^t)
\end{equation}
where $D$ is the degree matrix, $\tilde{A} = D^{-\frac{1}{2}} A D^{-\frac{1}{2}}$ is the degree normalized adjacency matrix, $W^t$ is a trainable weight matrix. Such aggregating operation can be stacked multiple times, to progressively take into account distantly connected nodes.

\subsection{Differentiable Pooling}

For particular tasks, e.g. graph classification, a hierarchical representation of the input graph in terms of the constituting nodes adopted for nodes cluster learning. Such operation is formally denoted as node clustering and can be iteratively applied in a similar fashion to message-passing to a graph to determine its building node groups. In the domain of neural networks, Differentiable Pooling (DiffPool)~\cite{ying2018diffpool} is the first example of node clustering in a differentiable way. In particular, DiffPool views node clustering as a soft node assignment task, where $n$ input nodes are associated with $k$ node clusters. Given input node embeddings $H \in \mathbbm{R}^{n \times d}$, the pooling layer is defined as follows:
\begin{equation} \label{eq:diffpool:pooling}
    P = softmax(H W_P)
\end{equation}
where $W_P \in \mathbbm{R}^{d \times k}$ is a learnable weight matrix and $P \in \mathbbm{R}^{n \times k}$ is the pooling soft assignment matrix. Lastly, node cluster embeddings $\tilde{H} \in \mathbbm{R}^{k \times d}$ are determined as follows:
\begin{equation} \label{eq:diffpool:new_nodes}
    \tilde{H} = P^T H
\end{equation}

Likewise, the adjacency matrix $\tilde{A} \in \mathbbm{R}^{k \times k}$ for new node clusters is built using previous node adjacency matrix $A$:
\begin{equation} \label{eq:diffpool:new_adjacency}
    \tilde{A} = P^T A P
\end{equation}

\subsection{Tree Kernels}

In the framework of kernel machines, a decision function $f(x)$ is defined as the weighted combination of the contributions of a subset of training elements, named support vectors:
\begin{equation}
    f(x) = \sum_{i=1}^m \alpha_i y_i K(x, x_i)
\end{equation}
The kernel function is a symmetric, positive-definite function that has to capture the degree of similarity between two examples. In the case of a TK, similarity is typically measured by taking into account the number of substructures that the two trees have in common. Such substructures are usually named fragments, and each fragment can be thought of as a feature upon which the similarity is evaluated. Thus, it is clear that different fragments induce different TK functions, and complex definitions of fragments can produce very rich feature sets. Given two trees $T_x$ and $T_z$, a TK is defined as:
\begin{equation}
K(T_x, T_z) = \sum_{n_x \in N_{T_x}} \sum_{n_z \in N_{T_z}} \Delta(n_x, n_z)
\end{equation}
where $N_{T_x}$ and $N_{T_z}$ the set of nodes in $T_x$ and $T_z$, respectively, and $\Delta(\cdot, \cdot)$ computes the similarity between two such nodes, by exploiting the chosen fragments.

For the sake of this paper, we will consider three of the most widely used TKs for NLP: namely, the SubTree Kernel (STK), the SubSet Tree Kernel (SSTK) and the Partial Tree Kernel (PTK). In STK, a fragment is any subtree of the original tree. SSTK is very similar to STK, but the subtree can terminate also at pre-terminals. PTK is much more general, since any portion of a subtree is a possible fragment. For all the technical details of the implementation of these kernels, see~\cite{moschitti2006making} and references therein

%% file: tcgnns.tex
\begin{figure*}[tb]
    \centering
    \includegraphics[width=0.9\linewidth]{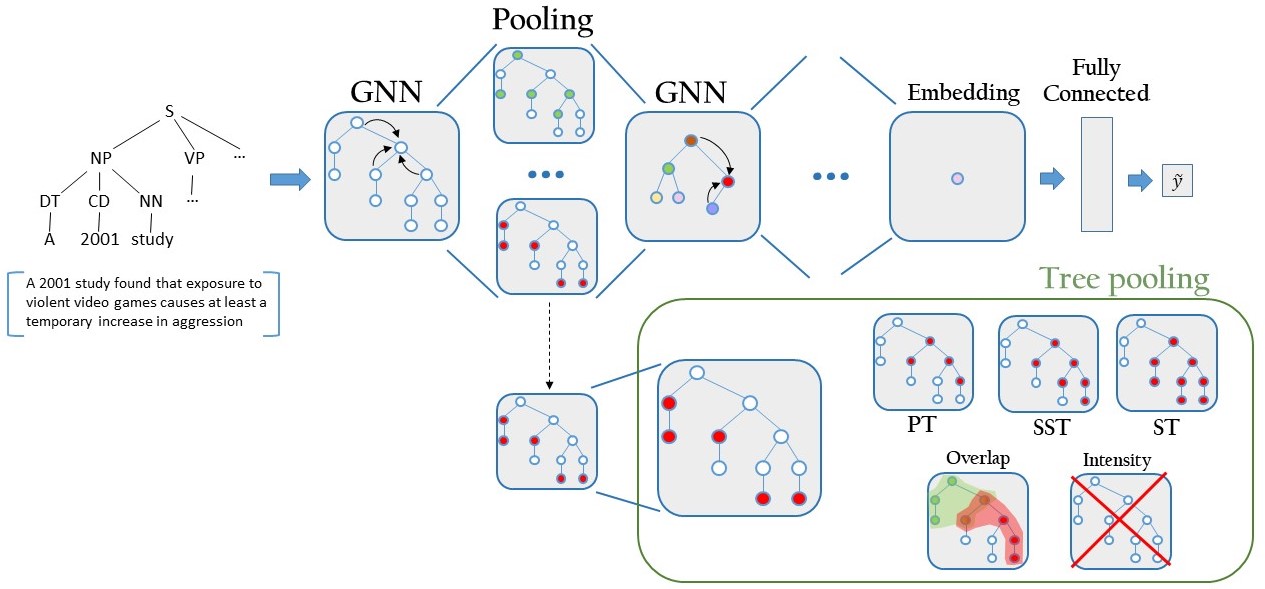}
    \caption{Tree-constrained GNN architecture. The constituency tree of a given sentence is fed as input to the network. The latter is defined as a stack of GCN and DiffPool layers, where the last DiffPool block has $k=1$ to determine a final tree embedding. The retrieved tree embedding is fed to a multi-layer perceptron for classification. In the case of tree regularizations, each pooling node cluster is subdued to such constraints for structured learning.}
    \label{fig:model_schema}
\end{figure*}

Inspired by the effectiveness of tree kernels, we investigate the problem of integrating such structured notion into deep learning models. Conversely to well-known tree compliant solutions, such as Recursive Neural Networks (RecNN)~\cite{sperduti1997supervised} and Tree-LSTM~\cite{tai2015}, we formulate the task of learning structured node clusters as the building block features of the input graph. In particular, the problem of extracting kernel specific tree fragments is viewed as a node soft assignment task by constraining the pooling operation. Each pooled node cluster should be compliant to a specific tree definition, dictated by a given tree kernel. 
Differently from node clustering, tree extraction does not necessarily impose mutually exclusive assignments. In most cases, node sharing between extracted trees is unavoidable due to the given fragment definition (e.g., for the STK).

\subsection{Naive Solution}
As a first proposal, we present a naive solution that determines all possible tree fragments of a given input tree before-hand. Subsequently, the pooling layer learns to extract the most relevant fragments for the task during training. Formally, we introduce a tree fragments matrix $T \in n \times s$ alike the pooling matrix $P$, where $s$ is the number of previously identified tree fragments. Subsequently, $P$ is constrained to match \textit{one and only one} column on $T$. Intuitively, such approach can only be considered when the number of nodes $n$ is limited. In particular, certain tree definitions like PTK and SSTK are of combinatorial complexity. Therefore, complete tree fragments enumeration cannot stand as a general purpose solution.

\subsection{Tree Structure Regularization}
Rather than exhaustively enumerating tree fragments, a more general approach consists in injecting each tree definition into the model itself. In this way, learning tree fragments is declared as a constrained differentiable operation that does not alter model architecture. It is then possible to automatically extract the most relevant fragments according to a specific kernel definition, without the need to exhaustively list all of them. Therefore, we only need to define appropriate regularizations that reflect the type of structured output we would like to obtain. Formally, we constrain Eq.~\ref{eq:diffpool:pooling} so that each node cluster minimizes all the constraints associated with the given fragments:
\begin{equation} \label{eq:constrained_pooling}
    P = softmax(H W_P) \quad s.t. \,\, \textbf{C}^{frag}(P) = 0
\end{equation}
where $\textbf{C}^{frag} = \{C_1^{frag}, \dots, C_\ell^{frag} \}$ is the set of constraint functions associated with a specific fragment definition. Formally, we label Tree-Constrained GNN (TC-GNN) a GNN model with a constrained differentiable pooling layer as in Eq.~\ref{eq:constrained_pooling}.

\subsection{Kernel-induced constraints}

In order to define appropriate tree constraints according to a given tree kernel, it is necessary to decompose the concept of tree into its constituting properties.
\begin{enumerate}
    \item {[\textit{Contiguity}]} A tree is a connected set of nodes.
    \item {[\textit{Root node}]} A tree has only one node at the lowest depth, denoted as the root node.
    \item {[\textit{No cycles}]} A tree allows no cycles.
    \item {[\textit{Root span}]} All nodes in a tree belong to the root node span, i.e. the set of children nodes, either directly or indirectly, of the root node.
\end{enumerate}

Additionally, we consider tree kernel-specific definitions:
\begin{enumerate}[label=\Alph*)]
    \item {[\textit{ST}]} for ST, all leaf nodes in the root node span must be considered;
    \item {[\textit{SST}]} for SST, all non-leaf nodes of a node $x$ in the root node span have to be considered if $x$ is considered.
\end{enumerate}

Note that in the case of tree structured inputs, the satisfaction of the \textit{Contiguity} property, inherently satisfies \textit{Root node}, \textit{No Cycles} and \textit{Root span} properties. Additionally, for PTK it is only required to satisfy the connected set of nodes properties in the case of tree structured inputs.

\subsubsection{Contiguity constraint}

Inspired by~\cite{bianchi2020spectral}, the constraint solely relies on the graph topology as follows:
\begin{equation} \label{eq:tree_rules:contiguous_constraint}
    C^{cont} = 1 - \frac{Tr \left (P^T (\overrightarrow{A} - diag(A)) P \right )}{Tr \left ( P^T P - \frac{1}{\mathds{1}_{P_i^2 \ge \epsilon}} P^T P \right )}
\end{equation}
where $\overrightarrow{A}$ is a masked version of $A$ that only considers forward connections, i.e., from root to leaves, and $\epsilon \ll 1$. The numerator computes node-to-node intensities, i.e., edges. Note that the numerator excludes self-intensities, i.e. the main diagonal is removed. On the other hand, the denominator just computes self-node intensities, minus the mean self-node intensity value. In simple terms, the constraint could be thought as the ratio between pooled edges and pooled nodes minus 1, since in a tree the number of edges is equal to $n - 1$, where $n$ is the number of nodes. Technically speaking, the mean self-node intensity could be replaced with a max as a more greedy estimate. Experimentally, a masked mean to address large input trees was chosen as a more stable solution. An additional side effect of this constraint is to ensure pooling intensity uniformity. Granting more flexibility in terms of pooling intensity is possible, but requires some sort of node intensity capping, which, however, may lead to harder optimization\footnote{The original formulation of tree constraints followed this perspective, but it was later substituted due to their difficulty of optimization.}.

\subsubsection{ST Constraint}

In order to be a ST, a connected set of nodes requires to consider all the nodes under the root node span. We can impose this requirement locally, by enforcing pooling children nodes of node $x$ when $x$ is pooled as well. The ST constraint leverages graph topology in terms of adjacency matrix $A$ and corresponding degree matrix $D$, which simply associates to a node its branching factor. We can obtain $D$ from $A$ by summing row-wise. Formally, the ST constraint is defined as follows:
\begin{equation}
    C^{ST} = 1 - \frac{Tr \left ( P^T (\overrightarrow{A} + \overleftarrow{A} (1 - L)) P \right )}{Tr \left ( P^T (\overrightarrow{D} + \overleftarrow{D}(1 - L)) P \right )}
\end{equation}
where $\overrightarrow{A}$ and $\overleftarrow{A}$ are masked versions of $A$ that only consider forward and backward connections, respectively. $L$ is the leaf mask vector, i.e., it is 1 for leaf nodes and 0 otherwise. It can be computed by summing $\overrightarrow{A}$ row wise (leaf nodes are the only nodes with no children). The numerator computes node-to-node intensities, i.e., edges, but masks some of them. More precisely, the first term of the numerator, $P^T \overrightarrow{A} P$ computes forward edges only. On the other hand, the second term, $P^T (\overleftarrow{A} (1 - L)) P$ computes backward edges only for leaf nodes. The denominator follows the same perspective, but it multiplies self-node intensities by corresponding node degree.

\subsubsection{SST Constraint}

The SST constraint can be considered as a special case of the ST constraint.
\begin{equation}
    C^{SST} = 1 - \frac{Tr \left ( P^T_L \overrightarrow{A} P_L \right )}{Tr \left ( P^T_L \overrightarrow{D}_L P_L \right )}
\end{equation}
where $P_L = PL$ and $D_L = \sum_{\{j, L_j \ne 0\}} A_{i, j}$. The main difference with ST is that in SST we are just filtering out leaf nodes. In particular, the numerator considers forward node-to-node intensities between non-leaf nodes, whereas the denominator multiplies non-leaf self-nodes intensities by their forward masked degree value (leaf nodes are not considered).

\subsection{Sigmoid-based Pooling}

Tree regularized pooling does not necessarily enforce mutually exclusive assignments. In other terms, a node might be assigned to multiple tree fragments. This scenario is particular common when considering tree definitions like ST and SST. On the other hand, differentiable pooling (Eq.~\ref{eq:diffpool:pooling}) applies a softmax operation, which might hinder constraint satisfaction, especially when considering large values of $k$. To this end, we replace softmax with a sigmoid activation function. By doing so, on the one hand we lose the softmax advantage of automatically building distinguishable node clusters. On the other hand, we obtain some desired properties. For instance, a low value of $k$ forces a softmax-based pooling to define large tree fragments. However, the sigmoid activation function allows to have a small amount of relevant sub-trees, while maintaining freedom of choice regarding their composition. Furthermore, sigmoid-based pooling potentially allows node filtering by not assigning a node to any node cluster.

\subsection{Avoiding degenerate scenarios}

The attained flexibility of tree constraints is inherently accompanied by the presence of degenerate scenarios due to the soft assignment operation. In particular, similarly to softmax-based pooling, we have to deal with two degenerate behaviours: i) all node clusters reflect the same tree structure; ii) nodes are assigned to a single node cluster while the remaining node clusters are ignored. Albeit there exist valid solutions as in~\cite{bianchi2020spectral}, some modifications are required since softmax properties no longer hold true.

\subsubsection{Overlap Constraint}

To overcome uniform node clusters, it is necessary to regularize the degree of overlap among node clusters. Differently from the orthogonality loss used in~\cite{bianchi2020spectral}, we allow tree fragments to overlap up to a controlled threshold, without enforcing node clusters to be of the same size. Thus, we devise the overlap constraint solely on the ratio between intra and self node clusters intensities as follows:
\begin{equation} \label{eq:tree-rules:overlap}
    C^{overlap} = \left \rVert \max \left ( \frac{\tilde{P}}{\sum_j diag(P^TP)_j} - \delta, 0 \right ) \right \rVert_F
\end{equation}
where $\tilde{P} = P^TP - diag(P^TP)$ and $\delta \in [0, 1]$ is the overlap threshold. In other terms, Eq.~\ref{eq:tree-rules:overlap} computes the intensities of shared nodes between pairs of node clusters, normalized by each node cluster self-intensity. Such derived ratio must not exceed the overlap threshold $\delta$. 

\subsubsection{Minimum Intensity Constraint}

The second degenerate scenario to consider regards empty node clusters. Similarly to the overlap constraint, we introduce a regularization term that penalizes node clusters with low intensity. Switching to sigmoid-based pooling loses the convenient softmax property of preserving the cumulative node intensity between input and the pooled graphs. To overcome this issue, we enforce a minimum node cluster intensity as follows:
\begin{equation} \label{eq:tree-rules:minimum-intensity}
    C^{intensity} = \left \rVert \max \left ((\alpha \frac{n}{k}) I - diag(P^T P), 0 \right ) \right \rVert_F
\end{equation}
where $\alpha \in [0, 1]$ is a hyper-parameter controlling the minimum intensity threshold. Such constraint enforces each un-normalized node cluster self-intensity to be greater than or equal to $\alpha \frac{n}{k}$. In other terms, each node cluster self-intensity should at least match the intensity of a node cluster of $\frac{n}{k}$ nodes whose self-intensity is equal to $\alpha$. Note that the presented constraint formulation is just one instance out of many solutions that works well for our case study. For instance, one could simply ignore $k$ and define a specific node cluster threshold.

\subsection{Adaptive Learning}

A major problem with learning soft tree constraints regards the nature of the learning problem. More precisely, the proposed tree regularizations are not always concurrent throughout training. This issue can lead to unwanted locally optimal solution where some tree constraints are highly violated. Since we explicitly require distinguishable tree fragments, such intermediate results should be avoided. A well-defined structured representation also increases model interpretability by directly inspecting the obtained node pooling. As an example of conflict between constraints, consider the overlap and minimum intensity constraints (Eqs.~\ref{eq:tree-rules:overlap} and~\ref{eq:tree-rules:minimum-intensity}). Both of them could be simply minimized by pooling sparse node groups. However, the contiguity constraint (Eq.~\ref{eq:tree_rules:contiguous_constraint}) would be largely violated as a consequence. Likewise, kernel specific constraints (ST and SST) cannot be satisfied if inter node clusters constraints prevail.

As a naive approach, one could try to calibrate ad-hoc weighting coefficients for constraints, in order to find their best linear combination. However, such antagonist dynamics among constraints might occur multiple times during the learning process and also depend on the model initialization. Thus, an approach based on fixed coefficients hardly leads to a solution where the majority of the constraints are satisfied. To this end, we propose to adopt the dual Lagrangian framework~\cite{FiorettoHM0B020} in order to define dynamic coefficients that are periodically updated during learning proportionally to the degree of constraints violation. Under this perspective, a constraint-based problem is relaxed as follows:
\begin{equation}
    \mathcal{L} = \argmin_\theta f_\theta(x) + \lambda g(x)
\end{equation}
where $\lambda \ge 0$ are the Lagrangian multipliers. In order to obtain the best Lagrangian relaxation, the Lagrangian dual can be used to find the best $\lambda$ value:
\begin{equation}
    \mathcal{L}^{dual} = \argmax_{\lambda \ge 0} \left ( \mathcal{L} \right )
\end{equation}

In our experimental scenario, the model is trained to optimize the following relaxed constrained objective:
\begin{equation}
    \mathcal{L} = \mathcal{L}^{CE} + \sum_i \lambda_i C^{frag}_i
\end{equation}
where $\mathcal{L}^{CE}$ is the cross-entropy loss for classification.

%% file: experiments.tex
As a testbed for our approach, we consider argumentative sentence detection, a sub-task of argument mining. The goal of argument mining is to automatically extract arguments for text collections. One of the major tasks is to identify argument components, such as claims or premises, within a sentence. The problem is formulated as a binary classification task where a sentence can either by labelled as a particular argument component or be evaluated as non-argumentative~\cite{lippi2016toit}. Depending on the annotation schema used in a given corpus, the task is slightly different, as it involves the identification of claims, premises, or both.
Several approaches have tried to shed light on the key properties of arguments, in order to determine common patterns, e.g., syntactic and rhetoric, among single or multiple domains~\cite{lippi2016toit,lawrence2020argument}. Primarily inspired by the work by Lippi and Torroni~\cite{lippi2015context}, we seek to extract tree-structured word patterns that in combination with automatically inferred semantic features may help in the identification of argumentative phrases. To ensure reproducibility, we made all our code and data available at the following repository:~\url{https://bitbucket.org/hl2exe/tcgnn/}.

\subsection{Data} \label{sec:data}

We carry out an extensive experimental evaluation on four distinct datasets for argumentative sentence detection.

\begin{itemize}
    \item \textbf{IBM2015~\cite{rinott2015show}}: an argumentative corpus built in the context of the Debater project~\cite{slonim2021autonomous}. The data set consists of a collection of Wikipedia pages, grouped into 58 topics. Annotations consist in context-dependent claims (i.e., with the underlying assumption that the topic is given) and evidence (i.e., premises) supporting such claims.
    
    \item \textbf{UKP Sentential Argument Mining Corpus (UKP-Sent)~\cite{stab-etal-2018-cross}}: it contains a large variety of heterogeneous sources, including news reports, editorials, blogs, debate forums and encyclopedia articles, on eight controversial topics. The argument annotation schema distinguishes arguments attacking or supporting a given stance, from sentences that are not argumentative.
    
    \item \textbf{Persuasive Essays Corpus (PE)~\cite{stab2017parsing}}: a collection of 402 documents extracted from an online community regarding essays discussion and advise. Documents are annotated at token-level and the argument annotation schema distinguishes the following argumentative components: major claim, claim, and premise.
    
    \item \textbf{PubMed RCT Abstracts (AbstRCT} \cite{Mayer2020TransformerBasedAM, Mayer2018ArgumentMO}: a collection of abstracts from scientific papers regarding randomized control trials for the treatment of specific diseases. The annotation schema reflects the well known claim-premise model~\cite{lippi2016toit} and is applied at token-level.
\end{itemize}

Table~\ref{table:dataset-info} provides a detailed summary of the corpora. In particular, since we are mainly interested in the detection of arguments at sentence-level, for token-level annotated datasets (such as PE and AbstRCT) we only consider argumentative component for classification~\cite{galassi-etal-2018-argumentative}.

\begin{table*}
\centering
\caption{Corpora for argumentative sentence detection.}
\label{table:dataset-info}
\begin{tabular}{cccc}
\multicolumn{1}{c}{\textbf{Corpus}} & \multicolumn{1}{c}{\textbf{No. Sentences}} & \multicolumn{1}{c}{\textbf{Class Distribution}} & \multicolumn{1}{c}{\textbf{Task}} \\
\hline
IBM2015         & 82,718 & \begin{tabular}[c]{@{}c@{}}2,388 claim,\\ 4,614 evidence,\\ 940 claim and evidence,\\76,656 non-argumentative\end{tabular} & \begin{tabular}[c]{@{}c@{}}Claim vs. non-claim,\\Evidence vs. non-evidence\\\end{tabular} \\
\hline
UKP-Sent        & 25,492 & \begin{tabular}[c]{@{}c@{}}6,195 argument against,\\ 4,944 argument for,\\14,353 non-argumentative\end{tabular} & \begin{tabular}[c]{@{}c@{}}Argument vs. non-argument\end{tabular} \\
\hline
PE              & 6,826 & \begin{tabular}[c]{@{}c@{}}751 major claims,\\ 1,506 claim,\\ 3,832 premise\end{tabular} & \begin{tabular}[c]{@{}c@{}}Merged claims vs. premise,\\\end{tabular} \\
\hline
AbstRCT         & 35,012 & \begin{tabular}[c]{@{}l@{}}1,390 claim,\\2,808 premise\end{tabular} & \begin{tabular}[c]{@{}l@{}}Claim vs. premise \end{tabular} \\
\end{tabular}
\end{table*}

\subsection{Experimental setup}
To ensure direct comparison with existing baselines, we follow, whenever possible, the same evaluation methodology employed in the literature. As mentioned in Section~\ref{sec:data}, for PE and AbstRCT we focus on argumentative component categorization in order to properly define a sentence-level argumentative classification. This simplification avoids a multi-label setting where phrases could contain multiple argumentative components. Such setting occurs for IBM2015 where, in some cases (see Table~\ref{table:dataset-info}), a sentence both contains a claim and an evidence. To solve this issue, as in~\cite{lippi2016margot} we carry out two distinct binary classification tasks concerning the identification of a single argumentative component at a time. Additionally, for PE the distinction between major claim and claim is mainly determined by their position in the essay. In particular, there are cases in which the same sentence is distinctly labelled belonging to both classes within the same paragraph. To better focus on the identification of unique structured features for each class, we merge the two claim classes into a single one. For UKP-Sent, we consider a binary classification task that discriminates between argumentative and non-argumentative sentences. The motivation lies on the employed annotation schema, which is entirely focused on the stance information (for/against), rather than on distinct argumentative components. In order to guarantee the same type of task, i.e., the identification of argumentative components, we merge labelled arguments into a single class.

As competitors, we consider the TK-SVM classifier with SSTK as in~\cite{lippi2016margot}, as well as LSTM-based models of increasing complexity: (i) a bi-directional LSTM that works on the input sentence (BLSTM); (ii) a bi-directional LSTM that sees the constituency tree as a sequence\footnote{Nodes are listed as a depth-first enumeration starting from the root node.} (Syn-BLSTM); (iii) a combination of (i) and (ii) via a self-attention layer (Dual-Syn-BLSTM). Additionally, we also consider standard tree-aware LSTM models like Tree-LSTM~\cite{tai2015}. At training time, all neural models are regularized by applying early stopping on the validation F1-score.
We hereby report additional evaluation details for each data set:
\begin{itemize}
    \item \textbf{IBM2015}: The original corpus is organized into two splits: 19 topics are used as a held-out set for model selection and hyperparameter tuning, whereas the remaining 39 topics are used for the final training and evaluation. We adopt a 5-fold cross-validation on the latter set of 39 topics, while we calibrate the hyper-parameters of our architecture using the available held-out set. 
    
    \item \textbf{UKP-Sent}: We consider the same leave-one-out (LOO) test, repeated 10 times, as in~\cite{reimers-etal-2019-classification}. We do not compare with strong LSTM-based and BERT-based baselines as reported in~\cite{reimers-etal-2019-classification} due to the different training setup. However, in early experiments with the same classification task as in~\cite{reimers-etal-2019-classification} the performance of GNN models nearly matched that of LSTM-BERT and BERT-base.
    
    \item \textbf{PE}: The original corpus is organized into two splits: 322 essays for training, and 80 for test. We performed a 5-fold cross-validation on the training data for model selection, and we finally evaluate performance on the test set. Due to the limited size of the corpus, for neural architectures we also employ a multi-start procedure, choosing the best model on the validation set via early stopping.
    
    \item \textbf{AbstRCT}: We follow the same evaluation methodology as in~\cite{galassi2021multitask}, where a repeated train-and-test procedure is considered both for tuning and testing. In addition to the neural baselines introduced for the other corpora, we consider state-of-the-art models for this task as reported in~\cite{galassi2021multitask}. In particular, these baselines are complex neural architectures mainly based on residual networks and the attention mechanism~\cite{Galassi_2020}.
\end{itemize}

\subsection{Results}

\textbf{IBM2015}: Table~\ref{table:results_IBM2015} reports classification performance for the IBM2015 corpus. 
In both cases, GNN models outperform their recurrent counterparts by a large margin. Such results corroborates the hypothesis that the information encoded by constituency trees is better handled by GNNs rather than recurrent models. The advantage is further emphasized when considering the longer input sequence defined by the constituency tree, with respect to the related sentence, which recurrent baselines struggle to handle properly.\footnote{The 99\% quantile concerning number of nodes per tree is 160.} The introduction of the pooling layer brings particular benefit to the task of claim detection. For both classification tasks, tree-regularized GNNs manage to consistently improve performance with respect to their standard pooling-based counterpart, while satisfying the imposed structured constraints. Such result is aligned with our initial hypothesis that arguments of the same class share common underlying linguistic patterns.

\textbf{UKP-Sent}: Table~\ref{table:results_UKP_merged} reports classification performance for UKP-Sent corpus. GNN models significantly outperform other neural baselines by a large margin. Additionally, tree-aware GNNs consistently improve over the standard and the pooling-based models. Such results, coupled with the notable performance of TK-SVM, which matches tree-GNN models, suggest that structured patterns are valuable indicators for detecting argumentative phrases.

\textbf{PE}: Table~\ref{table:results_essays_merged} reports classification performance for PE corpus. GNNs perform better than recurrent baselines. Differently to IBM2015, all models differ of 1-2 percentage points when considering the macro F1-score. Nonetheless, when considering the F1@C column, GNNs significantly outperform all the other models. In particular, tree-regularized GNNs achieve the highest performance for both argument classes.

\textbf{AbstRCT}: Table~\ref{table:abstRCT_merged} reports classification performance for AbstRCT corpus. As in \cite{galassi2021multitask}, we report classification metrics for each test set. GNN models manage to reach satisfying performance compared to ResArg and ResAttArg strong baselines. The GNN extension with DiffPool and, in particular, TC-GNNs consistently improve performance with respect to the base GNN. Considering recurrent baselines, the structured representation of the constituency tree is particularly useful for the BLSTM baselines, where both the Syn-BLSTM and the Dual-Syn-BLSTM nearly match the ResArg model.

\begin{table}
    \centering
    \caption{Classification performance on IBM2015 corpus. For each argumentative class, i.e. claim and evidence, we report achieved f1-score on the test set.}
    \label{table:results_IBM2015}
    \begin{tabular}{ccc}
    & F1@C & F1@E \\
    \hline
    BLSTM & 15.0 & 16.4 \\
    Syn-BLSTM & 14.0 & 14.7 \\
    Dual-Syn-BLSTM & 14.7 & 15.3 \\
    Tree-LSTM & 11.5 & 12.4 \\
    TK-SVM & 12.7 & 8.9 \\
    GNN & 18.5 & 18.8 \\
    GNN + DiffPool & 19.7 & 18.4 \\
    TC-GNN (STK) & 18.3 & 18.5 \\
    TC-GNN (SSTK) & \underline{20.1} & \textbf{19.2} \\
    TC-GNN (PTK) & \textbf{21.3} & \underline{19.1} \\
    \end{tabular}
\end{table}

\begin{table}
    %\advanced\leftskip-1.0cm
    \centering
    \caption{Classification performance on UKP-Sent corpus with merged argument classes: class Arg contains both pros and cons for each stance.}
    \label{table:results_UKP_merged}
    \begin{tabular}{ccccc}
    & F1@Arg \\
    \hline
    BLSTM & 0.513 \\
    Syn-BLSTM & 0.572 \\
    Dual-Syn-BLSTM & 0.544 \\
    Tree-LSTM & 0.517 \\
    TK-SVM & 0.605 \\
    GNN & 0.595 \\
    GNN + DiffPool & 0.590 \\
    TC-GNN (STK) & \textbf{0.610} \\
    TC-GNN (SSTK) & \textbf{0.610} \\
    TC-GNN (PTK) & \underline{0.608} \\
    \end{tabular}
\end{table}

\begin{table}
    \centering
    \caption{Classification performance on PE corpus with merged claim classes: class C contains both claims and major claims.}
    \label{table:results_essays_merged}
    \begin{tabular}{ccccc}
    & F1@C & F1@PR & F1@Macro \\
    \hline
    BLSTM & 62.05 & 78.16 & 70.10 \\
    Syn-BLSTM & 60.69 & 77.61 & 69.14 \\
    Dual-Syn-BLSTM & 59.79 & 79.10 & 69.44 \\
    Tree-LSTM & 54.78 & 57.48 & 56.13 \\
    TK-SVM & 62.11 & 77.24 & 69.68 \\
    GNN   & 64.17 & 78.07 & 71.12 \\
    GNN + DiffPool & 64.08 & 78.03 & 71.05 \\
    TC-GNN (STK) & 65.42 & 77.82 & \textbf{71.62} \\
    TC-GNN (SSTK) & 64.40 & 78.02 & 71.21 \\
    TC-GNN (PTK) & 64.36 & 78.74 & \underline{71.55} \\
    \end{tabular}
\end{table}

\begin{table*}
    \centering
    \caption{Classification performance on AbstRCT. We report per class F1-score and macro F1-score for each reported test set.}
    \label{table:abstRCT_merged}
    \begin{tabular}{cccc|ccc|ccc}
    & \multicolumn{3}{c|}{Neoplasm}       & \multicolumn{3}{c|}{Glaucoma}       & \multicolumn{3}{c}{Mixed}          \\
    \multicolumn{1}{l}{}         & F1@C & F1@PR & F1 & F1@C & F1@PR & F1 & F1@C & F1@PR & F1 \\
    \hline
    BLSTM                         & 78.80           & 88.61            & 83.71         & 77.04           & 91.04            & 84.04         & 77.39           & 89.33            & 83.36         \\
    Syn-BLSTM               & 82.74           & 90.53            & \textbf{86.63}         & 79.37           & 91.43            & 85.40         & 82.81           & 91.15            & \underline{86.98}         \\
    Dual-Syn-BLSTM           & 81.58           & 89.70            & 85.64         & 79.86           & 91.31            & \underline{85.59}        & 82.32           & 90.81            & 86.56         \\
    TK-SVM  & 79.25           & 90.31            & 84.78         & 71.01           & 90.02            & 80.52         & 77.62           & 90.04            & 83.83         \\
    Tree-LSTM & 71.79 & 76.67 & 74.23 & 72.22 & 82.88 & 77.55 & 77.09 & 83.17 & 80.13 \\
    %Tree-Net &  42.30 & 26.11 & 34.21 & 38.35 & 28.36 & 33.36 & 41.69 & 26.37 & 34.03 \\ \hline
    ResArg                             & 82.04           & 90.31            & 86.18         & 79.48           & 91.59            & 85.53         & 82.35           & 91.13            & 86.74         \\
    ResAttArg                             & 82.27           & 90.11            & 86.19         & 80.72           & 91.79            & \textbf{86.26}         & 83.74           & 91.27            & \textbf{87.51}         \\
    GNN                  & 80.72           & 89.97            & 85.34         & 77.14           & 90.98            & 84.06         & 80.78           & 90.56            & 85.67         \\
    GNN + DiffPool             & 81.17          & 90.06            & 85.61         & 77.09           & 91.03            & 84.06         & 81.25           & 90.58            & 85.91         \\
    TC-GNN (STK)  & 82.35           & 90.50            & \underline{86.43}         & 78.10           & 91.16            & 84.63         & 82.45           & 91.07            & 86.76     \\ TC-GNN (SSTK) & 80.97           & 89.90            & 85.44         & 77.25           & 91.06            & 84.15         & 81.24           & 90.60            & 85.92         \\
    TC-GNN (PTK)  & 81.90          & 90.30            & 86.10         & 77.17           & 91.00            & 84.08         & 81.74           & 90.70            & 86.22         \\
\end{tabular}
\end{table*}

\subsection{Discussion}

\begin{figure*}[tb]
\begin{center}
\begin{scriptsize}
    \centering
\begin{tabular}{C{1.5cm}|C{4cm}|C{4cm}|C{4cm}}
& STK & SSTK & PTK \\
~ & ~ & ~ & \\
\hline
~ & ~ & ~ & \\
Claims
&
      \begin{tikzpicture}[level distance=20pt,sibling distance=12pt]
        \Tree[.S [.NP ][.VP [.MD should ][.VP ] ] ] 
      \end{tikzpicture}
&
      \begin{tikzpicture}[level distance=20pt,sibling distance=12pt]
      \Tree[.VP [.MD should ][.VP [.VB be ][.VP [.VBN ] ] ] ]
      \end{tikzpicture}
&
      \begin{tikzpicture}[level distance=20pt,sibling distance=12pt]
      \Tree[.NP [.ADJP [.JJ ][.JJ ] ][.NN ][.PP [.NP [.NNS ] ] ] ]
      \end{tikzpicture}
\\
~ & ~ & ~ & \\
\hline
~ & ~ & ~ & \\
Premises
&
      \begin{tikzpicture}[level distance=20pt,sibling distance=12pt]
      \Tree[.VP [.VB understand ][.NP [.NNS topics ] ] ]
      \end{tikzpicture}
&
      \begin{tikzpicture}[level distance=20pt,sibling distance=12pt]
      \Tree[.NP [.PRP\$ their ][.NNS studies ] ]
      \end{tikzpicture}
&
      \begin{tikzpicture}[level distance=20pt,sibling distance=12pt]
        \Tree[.NP [.NML [.JJ ][.NN ] ][.NN ] ]
      \end{tikzpicture}
\end{tabular}
    \caption{Some of the most frequent class-specific tree fragments extracted from TC-GNN models on the PE corpus (top: claims, bottom: premises).}
    \label{fig:fragments}
\end{scriptsize}
\end{center}
\end{figure*}
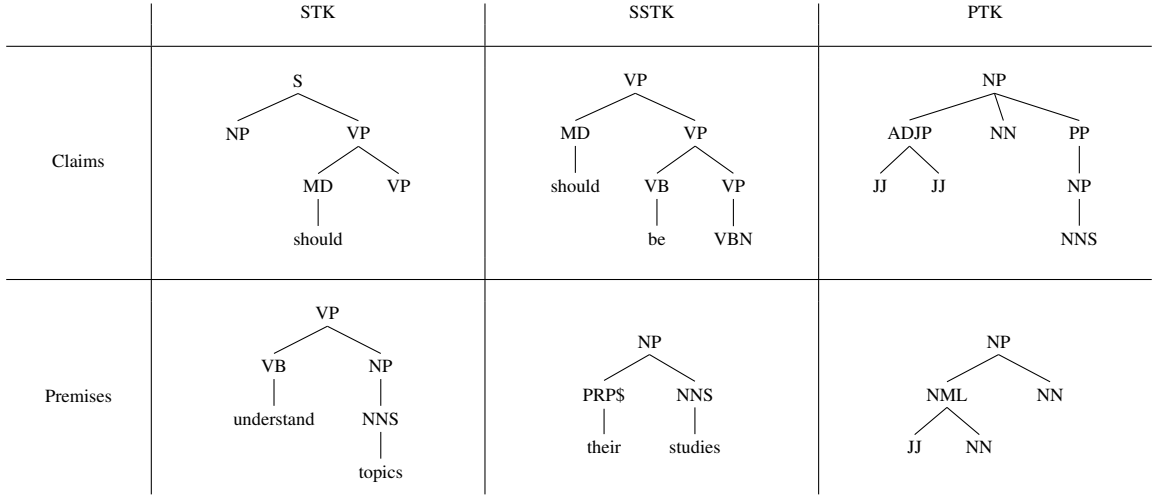

The positive results described so far corroborate our initial hypothesis that structured linguistic patterns are significant for the identification of arguments in a sentence. In particular, TC-GNNs reach increased generalization capabilities over the non-regularized counterpart model, while satisfying tree constraints to a certain degree. We hereby discuss how the GNN models are capable to minimize the imposed regularizations, whether recognizable structured patterns can be extracted for each argumentative class, and the overall introduced overhead.

First of all, we start by inspecting the difficulty of extracting tree fragments by regularizing the pooling layer of the GNN architecture. To do so, we compare the dual Lagrangian methodology with a fixed coefficient baseline, i.e., the $\lambda$ value for each constraint is initialized to a certain value and kept fixed throughout training. Table~\ref{table:satisfaction_IBM2015} reports classification performance and the average constraint value (the lower the better) for claim detection in the IBM2015 corpus. We can clearly see, how the use of the dual Lagrangian method is crucial both to reach higher performance and to achieve lower constraints violation. Moreover, an additional advantage of the approach relies on its simplicity and scalability. Fixed coefficients baseline require an extensive calibration phase in order to reach satisfying performance.
In particular, most of the evaluated configurations failed to reach significant performance. On the other hand, even by initializing all $\lambda$ values to 0, the dual Lagrangian method enforces constraints alignment and reduces the occurrence of degenerate scenarios where one of them is significantly violated. Further calibration of initial values might improve performance. We leave this investigation as future work. Lastly, in the case of a deeper model with stacked pooling layers, the dual Lagrangian method scales linearly with the number of constraints and of pooling layers\footnote{We consider a distinct set of multipliers for each pooling layer.}.

From a qualitative point of view, Figure~\ref{fig:fragments} reports some of the most frequent tree fragments for each TC-GNN model on the PE corpus. In particular, we consider fragments that are unique to a certain argumentative class, i.e., are not shared with other classes. By doing so, we are able to inspect structured patterns that are valuable indicators of an argumentative component, e.g., a claim or a premise. Due to pooling soft assignment, we apply an activation threshold in order to discriminate `active' nodes. We set the activation threshold to 0.3 in order to include a large pool of fragments.
Considering PTK fragments, we can see how the model focuses more on part-of-speech tags, i.e., lower levels of the constituency tree.
On the other hand, for SSTK and STK, the majority of fragments is centred near leaf nodes to satisfy kernel specific constraints. In both cases, claims are mainly identified by expressions with modal verbs, such as `should', whereas premises are often characterized by noun phrases.

Lastly, we also inspect the added overhead of described tree regularizations with respect to the base GNN model. Table \ref{table:timings} reports average training and inference times for all pooling based GNNs on the IBM2015 corpus. We considered the IBM2015 corpus as it is the one with the highest number of samples, thus, representing a valuable benchmark setting. Compared to the GNN + DiffPool model, TC-GNNs have comparable training and inference times, meaning that the introduced regularization overhead is negligible.

\begin{table}
\centering
\caption{Training and inference times on the IBM2015 corpus. For training, we report the average batch time in seconds, computed over 50 epochs. Standard deviation is in brackets.}
\label{table:timings}
\begin{tabular}{ccc}
              & Training      & Inference     \\
\hline
GNN + DiffPool & 0.065 (0.004) & 0.046 (0.032) \\
TC-GNN (STK)   & 0.075 (0.008) & 0.048 (0.077) \\
TC-GNN (SSTK)  & 0.119 (0.010) & 0.042 (0.061) \\
TC-GNN (PTK)   & 0.118 (0.009) & 0.060 (0.105) \\
\end{tabular}
\end{table}

\begin{table}
    \centering
    \caption{Constraint satisfaction for claim detection on the IBM2015 corpus. We report classification performance and the average constraint violation (the lower the better). Standard deviation is reported in brackets.}
    \label{table:satisfaction_IBM2015}
    \begin{tabular}{cccc}
    Kernel & Weights & F1@C & $C^{frag}$ \\
    \hline
    STK & Fixed & 11.8 & 0.62 (0.01) \\
    STK & Lagrangian & 18.3 & \textbf{0.19} (0.01) \\
    SSTK & Fixed & 17.2 & 0.70 (0.04) \\ %\hline
    SSTK & Lagrangian & 20.1 & \textbf{0.22} (0.02) \\
    PTK & Fixed & 15.2 & 0.54 (0.03) \\ %\hline
    PTK & Lagrangian & 21.3 & \textbf{0.50} (0.28) \\
    \end{tabular}
\end{table}

%% file: related.tex
\textbf{Graph Pooling} GNNs represent an emerging trend in AI due to their natural inclination for learning from structured data.
A recent research direction in the domain of graph classification regards learning valuable node clusters in the pooling stage, i.e., hierarchical learning. This process can be informally thought as decomposing a graph into a set of distinct node groups, which are viewed as its building blocks. The first example of hierarchical learning is given in~\cite{ying2018diffpool}, where a differentiable node clustering layer, labelled as DiffPool, defines a high-level graph representation via a soft node assignment operation. Subsequently, several node clustering architectures have been proposed in order to learn efficient node clusters while avoiding degenerate scenarios, such as uniform node clusters~\cite{bianchi2020spectral} or smooth node representations~\cite{mesquita2020rethinking}. Notable examples are given by SAGPool~\cite{lee2019selfattention}, minCutPool~\cite{bianchi2020spectral}, HGP-SL~\cite{zhang2019hierarchical} and SOPool~\cite{Wang_2020}. Differently from these approaches, we propose a simple solution for structured learning subdued to domain specific constraints, by applying regularizations at the pooling layer level. In particular, we do not require any modification of the model architecture, provided that node pooling is defined as a soft assignment operation. In this way, it is possible to encode high-demanding requirements by defining structured regularizations in a plug-and-play fashion.
Clearly, it is also possible to directly encode the described regularization functions within the graph convolution operation, by defining more powerful graph architectures such as in~\cite{morris2020weisfeiler, xu2019power}. In particular, our method defines a soft approximation of a TK, that is, the network learns to extract the $k$ most important tree fragments. A graph convolution based formulation would consider all tree fragments in the node encoding operation. In this case, the aggregation function (Eq.~\ref{eq:gcn:message-passing}) views node groups as neighbours in compliance with a given TK. To ensure efficiency, a hierarchical formulation as in~\cite{morris2020weisfeiler} might be considered. Yet, such method directly affects the network architecture and its formulation might be more complex than a set of soft regularization functions. On the other hand, the soft nature of such constraints inherently gives space to degenerate scenarios that, in order to be avoided, might require additional penalties and dynamic optimization methods.

% Tree LSTMs
\textbf{Structured Recurrent Models} Recurrent models have been widely adopted to tree data by adapting the recurrent cell update to the given structure. The most notable example in this setting is given by Recursive Neural Networks~\cite{sperduti1997supervised}, in which the recurrent cell state is updated recursively by navigating the tree structure bottom-up, i.e. from leaves to root. Subsequently, several model variants have been proposed in order to define more efficient and scalable state update methods. The most notable example is given by Tree-LSTM\cite{tai2015}. The main drawback of such models is that of handling binary tree structures only. Such limitation drastically impacts on the nature of the input\footnote{It is always possible to retrieve an equivalent binary tree in terms of node ordering. However, the tree structure significantly changes.}. To this end, several solutions, such as Tree-Net~\cite{chen2018treetotree}, ARTree-LSTM~\cite{XU2021115182} and CP-LSTM~\cite{castellana2020learning}, propose novel state update mechanisms that extend the model to any n-ary tree structure. Albeit our experimental scenario focuses on tree structured data only, our regularization methodology can be extended to any kind of domain (with ad hoc constraints) and to graph structures as well. For instance, we can consider TC-GNNs for graphs while imposing node clusters to define a DAG, similarly to~\cite{yu2019daggnn}. On the other hand, the aforementioned recurrent models are entirely centred on tree structured data. Regarding applications of GNNs to learn tree structured data, the closest examples to our experimental setting are~\cite{qiao2020tree} and~\cite{talak2021neural}. In the first, authors apply tree structured templates for node aggregation by exploiting domain specific entity relations. Such approach could be potentially be implemented as a tree regularized pooling operation where node clusters are enforced to match domain specific templates. Conversely, in~\cite{talak2021neural} a high level tree structure, denoted as H-tree is defined from a starting graph by applying the junction tree decomposition algorithm~\cite{jensen2013optimal} in order to achieve efficient message-passing. Such method is particularly important when handling large complex graphs. On the other hand, we are mainly interested in extracting few key structured features from input data for efficient and sparse tree representation and, potentially, tree comparison.

%% file: conclusions.tex
We presented a novel architecture for GNNs that exploits regularization terms that impose constraints on the node soft assignments function induced by the pooling mechanism. Such constraints are inspired by the notion of fragments that is used by TKs, that have achieved remarkable results in many applications, and especially in NLP.

We considered sentence classification in argument mining as a challenging case study for our approach, and we showed how the proposed architecture outperforms both other GNN-based systems and other strong competitors.

In the future, we plan to extend the work in several directions. First of all, we aim to experiment our architecture with multiple stacked layers, and we will consider some alternatives to using the GCN building block. Second, we plain to extend our methodology also to handling graphs: this would additionally require enforcing the adjacency matrix $A$ to be a DAG~\cite{yu2019daggnn}. Such methodology can either be applied to the input graph or to each pooling cluster. Finally, we will test the approach also to other domains, such as bioinformatics, or transportation networks.

%% file: appendix.tex
We choose GCN~\cite{KipfW16} as the base GNN within our architecture. Nonetheless, we consider few simple modifications to explore sufficiently complex model configurations. The GCN layer is mainly described via two distinct operations: (i) aggregation via message passing and (ii) node encoding update. Our implementation follows a variant of Eq.~\ref{eq:gcn:message-passing}:

\begin{equation}
    H^{t+1} = RNN_{cell}(MLP(A, H^t; \theta^{t+1}), H^t)
\end{equation}

that is, multiple non-linearities might be applied in sequence instead of a single one. In the case of a single layer, we go back to Eq.~\ref{eq:gcn:message-passing}. Lastly, a recurrent cell is applied to get the updated node encoding $H^{t+1}$, where the state is defined by the previous node encoding $H^t$.

In the case of differentiable pooling, the GCN layer is enhanced with the DiffPool layer as in Eqs.~\ref{eq:diffpool:pooling},~\ref{eq:diffpool:new_nodes},~\ref{eq:diffpool:new_adjacency}. Both the GCN and DiffPool layers are viewed as a single block (see Figure~\ref{fig:model_schema}) that can be stacked multiple times. Also in this case, we slightly modify aforementioned equations by considering MLPs instead of a single matrix multiplication as follows:
\begin{gather}
    P = softmax(MLP(H)) \\
    \tilde{H} = P^T MLP(H)
\end{gather}
that is, we transform current node encoding by applying a sequence of fully connected layers. In the case of determining $P$, the last layer is enforced to have $k$ neurons.

For argument classification, we label GNN a GCN layer with a DiffPool layer with $k = 1$ for graph embedding, while we label GNN + DiffPool the GNN model with an intermediate GCN and DiffPool layers with arbitrary $k$. Lastly, the graph embedding is fed to a multi-layer perceptron (MLP) for classification. Regarding hyper-parameter tuning, we calibrate each part separately, in the following order: (1) GCN layers for the GNN model; (2) the number of layers and neurons in each layer for the final MLP; (3) the intermediate GCN and DiffPool blocks; (4) regularization hyper-parameters $\alpha$ and $\delta$.